\pdfoutput=1
\documentclass[article,shortnames]{jss}

\newcommand*\pct{\scalebox{.9}{\%}}

\author{Marvin N. Wright\\Universit{\"a}t zu L{\"u}beck \And
        Andreas Ziegler\\Universit{\"a}t zu L{\"u}beck,\\University of KwaZulu-Natal}
\title{\pkg{ranger}: A Fast Implementation of Random Forests for High Dimensional Data in \proglang{C++} and \proglang{R}}

\Plainauthor{Marvin N. Wright, Andreas Ziegler}
\Plaintitle{ranger: A Fast Implementation of Random Forests for High Dimensional Data in C++ and R} 
\Shorttitle{\pkg{ranger}: Fast Random Forests in \proglang{C++} and \proglang{R}} 

\Abstract{
  We introduce the \proglang{C++} application and \proglang{R} package \pkg{ranger}. The software is a fast implementation of random forests for high dimensional data. Ensembles of classification, regression and survival trees are supported. We describe the implementation, provide examples, validate the package with a reference implementation, and compare runtime and memory usage with other implementations. The new software proves to scale best with the number of features, samples, trees, and features tried for splitting. Finally, we show that \pkg{ranger} is the fastest and most memory efficient implementation of random forests to analyze data on the scale of a genome-wide association study.
}
  
\Keywords{\proglang{C++}, classification, machine learning, \proglang{R}, random forests, \pkg{Rcpp}, recursive partitioning, survival analysis}
\Plainkeywords{C++, classification, machine learning, R, random forests, Rcpp, recursive partitioning, survival analysis}

\Address{
  Andreas Ziegler\\
  Institut f{\"u}r Medizinische Biometrie und Statistik \textit{and} Zentrum f{\"u}r Klinische Studien\\
  Universit{\"a}t zu L{\"u}beck\\
  Universit{\"a}tsklinikum Schleswig-Holstein, Campus L{\"u}beck\\
  Ratzeburger Allee 160 \\
  23562 L{\"u}beck, Germany\\
  \textit{and}\\
  School of Mathematics, Statistics and Computer Science\\
  University of KwaZulu-Natal\\
  Durban, South Africa

  Telephone: +49/451/500-2780\\
  Fax: +49/451/500-2999\\
  E-mail: \email{ziegler@imbs.uni-luebeck.de}
}

\usepackage{mdframed,xcolor}

\begin{document}
  
\begin{mdframed}[backgroundcolor=blue!20] 
\noindent This paper was downloaded from arXiv. It was published in final form in the Journal of Statistical Software as: \\[1mm]
Wright, M. N. \& Ziegler, A. (2017). ranger: A fast implementation of random forests for high dimensional data in C++ and R. Journal of Statistical Software 77:1-17. \\[1mm]
The final version can be downloaded for free at the JSS website: \\ \url{http://dx.doi.org/10.18637/jss.v077.i01}.
\end{mdframed}

\section{Introduction}

Random forests \citep[RF;][]{breiman2001} are widely used in applications, such as gene expression analysis, protein-protein interactions, identification of biological sequences, genome-wide association studies, credit scoring or image processing \citep{bosch2007,kruppa2013,qi2012}. Random forests have been described in several review articles, and a recent one with links to other review papers has been provided by \cite{ziegler2014}.

All of the available RF implementations have their own strengths and weaknesses. The original implementation by \cite{breiman2004} was written in \proglang{Fortran 77} and has to be recompiled whenever the data or any parameter is changed. The \proglang{R} implementation \pkg{randomForest} by \cite{liaw2002} is feature-rich and widely used. However, it has not been optimized for the use with high dimensional data \citep{schwarz2010}. This also applies to other implementations, such as \pkg{Willows} \citep{zhang2009} which has been optimized for large sample size but not for a large number of features, also termed independent variables. The \proglang{R} package \pkg{party} \citep{hothorn2006} offers a general framework for recursive partitioning, includes an RF implementation, but it shares the weaknesses of \pkg{randomForest} and \pkg{Willows}. The package \pkg{randomForestSRC} supports classification, regression and survival \citep{ishwaran2015}, and we study this package in greater detail below (Section~\ref{sec:runtime}). With \pkg{bigrf} \citep{lim2014} RF on very large datasets can be grown by the use of disk caching and multithreading, but only classification forests are implemented. The commercial implementation of RF is \pkg{RandomForests} \citep{salfordsystems2013}, which is based on the original \proglang{Fortran} code by Breiman with a license for commercial use. Here, licensing costs and closed source code limit the usability. A recent implementation available in \proglang{R} is the \pkg{Rborist} package \citep{seligman2015}. This package is studied in greater detail in Section~\ref{sec:runtime}. Finally, an RF implementation optimized for analyzing high dimensional data is \pkg{Random Jungle} \citep{schwarz2010, kruppa2014appl}. This package is only available as \proglang{C++} application with library dependencies, and it is not portable to \proglang{R} or another statistical programming language. 

We therefore implemented a new software package ``RANdom forest GEneRator'' (\pkg{ranger}), which is also available as \proglang{R} package under the same name. Our primary aim was to develop a platform independent and modular framework for the analysis of high dimensional data with RF. Second, the package should be simple to use and available in \proglang{R} with a computational performance not worse than \pkg{Random Jungle} 2.1 \citep{kruppa2014appl}. Here, we describe the new implementation, explain its usage and provide examples. Furthermore, we validate the implementation with the \pkg{randomForest} \proglang{R} package and compare runtime and memory usage with existing software.

\section{Implementation}
\label{sec:implementation}

The core of \pkg{ranger} is implemented in \proglang{C++} and uses standard libraries only. A standalone \proglang{C++} version can therefore easily be installed on any up-to-date platform. In the implementation, we made extensive use of \proglang{C++11} features. For example, parallel processing is available on all platforms with the \code{thread} library, and the \code{random} library is used for random number generation. 

The \proglang{R} package \pkg{Rcpp} \citep{eddelbuettel2011} was employed to make the new implementation available as \proglang{R} package, reducing the installation to a single command and simplifying its usage, see Section~\ref{sec:examples} for details. All data can be kept in \proglang{R}, and files do not have to be handled externally. Most of the features of the \pkg{randomForest} package are available, and new ones were added. For example, in addition to classification and regression trees, survival trees are now supported. Furthermore, class probabilities for multi-class classification problems can be estimated as described by \cite{kruppa2014theory, kruppa2014appl}. To improve the analysis of data from genome-wide association studies (GWAS), objects from the \proglang{R} package \pkg{GenABEL} \citep{aulchenko2007} can be directly loaded and analyzed in place. Genotypes are stored in a memory efficient binary format, as inspired by \cite{aulchenko2007}; see Section~\ref{sec:runtime} for the effect on memory usage. Genetic data can be mixed with dichotomous, categorical and continuous data, e.g., to analyze clinical variables together with genotypes from a GWAS. For other data formats, the user can choose between modes optimized for runtime or memory efficiency.

Implemented split criteria are the decrease of node impurity for classification and regression RF, and the log-rank test \citep{ishwaran2008} for survival RF. Node impurity is measured with the Gini index for classification trees and with the estimated response variance for regression trees. For probability estimation, trees are grown as regression trees; for a description of the concept, see \cite{malley2012}. Variable importance can be determined with the decrease of node impurity or with permutation. The permutation importance can be obtained unnormalized, as recommended by \cite{nicodemus2010}, or scaled by their standard errors \citep{breiman2001}. The prediction error is obtained from the out-of-bag data as the missclassification frequency, the mean square error, or as one minus the $C$ index \citep{harrell1982} for classification, regression, and survival, respectively.

We optimized \pkg{ranger} for high dimensional data by extensive runtime and memory profiling. For different types of input data, we identified bottlenecks and optimized the relevant algorithms. The most crucial part is the node splitting, where all values of all \code{mtry} candidate features need to be evaluated as splitting candidates. Two different splitting algorithms are used: The first one sorts the feature values beforehand and accesses them by their index. In the second algorithm, the raw values are retrieved and sorted while splitting. In the runtime-optimized mode, the first version is used in large nodes and the second one in small nodes. In the memory efficient mode, only the second algorithm is used. For GWAS data, the features are coded with the minor allele count, i.e., 0, 1 or 2. The splitting candidate values are therefore equivalent to their indices, and the first method can be used without sorting. Another bottleneck for many features and high \code{mtry} values was drawing the \code{mtry} candidate splitting features in each node. Here, major improvements were achieved by using the algorithm by \citet[p. 137]{knuth1985} for sampling without replacement. Memory efficiency has been generally achieved by avoiding copies of the original data, saving node information in simple data structures and freeing memory early, where possible. 

\pkg{ranger} is open source software released under the GNU GPL-3 license. The implementation is modular, new tree types, splitting criteria, or other features can be easily added by us, or they may be contributed by other developers.

\section{Usage and examples}
\label{sec:examples}

The \pkg{ranger} \proglang{R} package has two major functions: \code{ranger()} and \code{predict()}. \code{ranger()} is used to grow a forest, and \code{predict()} predicts responses for new datasets. The usage of both is as one would expect in \proglang{R}: Models are described with the \code{formula} interface, and datasets are saved as a \code{data.frame}. As a first example, a classification RF is grown with default settings on the iris dataset \citep{r2014}:

\begin{CodeChunk}
  \begin{CodeInput}
R> library("ranger")
R> ranger(Species ~ ., data = iris)
  \end{CodeInput}
\end{CodeChunk}

This will result in the output

\begin{CodeChunk}
  \begin{CodeOutput}
Ranger result
 
Call:
 ranger(Species ~ ., data = iris) 

Type:                             Classification 
Number of trees:                  500 
Sample size:                      150 
Number of independent variables:  4 
Mtry:                             2 
Target node size:                 1 
Variable importance mode:         none 
OOB prediction error:             4.00 %
  \end{CodeOutput}
\end{CodeChunk}

In practice, it is important to check if all settings are set as expected. For large datasets, we recommend starting with a dry run with very few trees, probably even using a subset of the data only. If the RF model is correctly set up, all options are set and the output formally is as expected, the real analysis might be run.

In the next example, to obtain permutation variable importance measures, we need to set the corresponding option:

\begin{CodeChunk}
  \begin{CodeInput}
R> rf <- ranger(Species ~ ., data = iris, importance = "permutation")
R> importance(rf)
  \end{CodeInput}
  \begin{CodeOutput}
Sepal.Length  Sepal.Width Petal.Length  Petal.Width 
 0.037522031  0.008655531  0.340307358  0.241499411
  \end{CodeOutput}
\end{CodeChunk} 

Next, we divide the dataset in training data and test data

\begin{CodeChunk}
  \begin{CodeInput}
R> train.idx <- sample(x = 150, size = 100)
R> iris.train <- iris[train.idx, ]
R> iris.test <- iris[-train.idx, ]
  \end{CodeInput}
\end{CodeChunk}

grow a RF on the training data and store the RF:

\begin{CodeChunk}
  \begin{CodeInput}
R> rf <- ranger(Species ~ ., data = iris.train, write.forest = TRUE)
  \end{CodeInput}
\end{CodeChunk}

To predict the species of the test data with the grown RF and to compare the predictions with the real classes, we use the code:

\begin{CodeChunk}
  \begin{CodeInput}
R> pred <- predict(rf, data = iris.test)
R> table(iris.test$Species, predictions(pred))
  \end{CodeInput}
  \begin{CodeOutput}
           setosa versicolor virginica
setosa         19          0         0
versicolor      0         11         2
virginica       0          1        17
  \end{CodeOutput}
\end{CodeChunk} 

For regression RF and survival RF, the interface is similar. The tree type is determined by the type of the dependent variable. For factors, classification trees are grown within an RF, for numeric values regression trees and for survival objects survival trees are used in the RF. If, for example, a dichotomous endpoint is saved as 0/1 in numeric format, it needs to be converted to a factor for classification RF. For survival forests, the \code{Surv()} function of the \pkg{survival} package \citep{therneau2000} is used to create a survival object, e.g., by

\begin{CodeChunk}
  \begin{CodeInput}
R> library("survival")
R> rf <- ranger(Surv(time, status) ~ ., data = veteran)
  \end{CodeInput}
\end{CodeChunk}

Next, we plot the estimated survival function for the first individual (not shown)

\begin{CodeChunk}
  \begin{CodeInput}
R> plot(timepoints(rf), predictions(rf)[1, ])
  \end{CodeInput}
\end{CodeChunk} 

Finally, we use \pkg{GenABEL} to read \pkg{Plink} \citep{purcell2007} files into \proglang{R} and grow an RF directly with this genetic data

\begin{CodeChunk}
  \begin{CodeInput}
R> library("GenABEL")
R> convert.snp.ped("data.ped", "data.map", "data.raw")
R> dat.gwaa <- load.gwaa.data("data.pheno", "data.raw")
R> phdata(dat.gwaa)$trait <- factor(phdata(dat.gwaa)$trait)
R> ranger(trait ~ ., data = dat.gwaa)
  \end{CodeInput}
\end{CodeChunk}

To use the \proglang{C++} version, \pkg{ranger} has to be compiled, or an executable binary version needs to be installed; for instructions, see the included \code{README} file. The interface is different from that in \proglang{R}, and datasets have to be available as ASCII files. The equivalent for the first classification example would be

\begin{CodeChunk}
  \begin{CodeInput}
> ranger --verbose --file iris.dat --depvarname Species --treetype 1
  \end{CodeInput}
\end{CodeChunk}

The command line output is similar to the \proglang{R} version. Further results are written to the files \code{ranger_out.confusion} and \code{ranger_out.importance}. For prediction, the estimated RF has to be saved to a file and loaded again:

\begin{CodeChunk}
  \begin{CodeInput}
> ranger --verbose --file iris_train.dat --depvarname Species
    --treetype 1 --write
> ranger --verbose --file iris_test.dat --predict ranger_out.forest
  \end{CodeInput}
\end{CodeChunk}

The predictions are written to a file called \code{ranger_out.prediction}.

Computational speed and memory usage are equal in the \proglang{R} and \proglang{C++} versions. The extreme memory efficient storage of GWAS data described in Section~\ref{sec:implementation} is, however, only available in the \proglang{R} version. We generally advise to use the \proglang{R} version.

\section{Validation}
To evaluate the validity of the new implementation, the out-of-bag prediction error and variable importance results were compared with the \proglang{R} package \pkg{randomForest}. Identical settings were used for both packages.

\begin{figure}[htbp]
  \centering
  \includegraphics[width=1\textwidth]{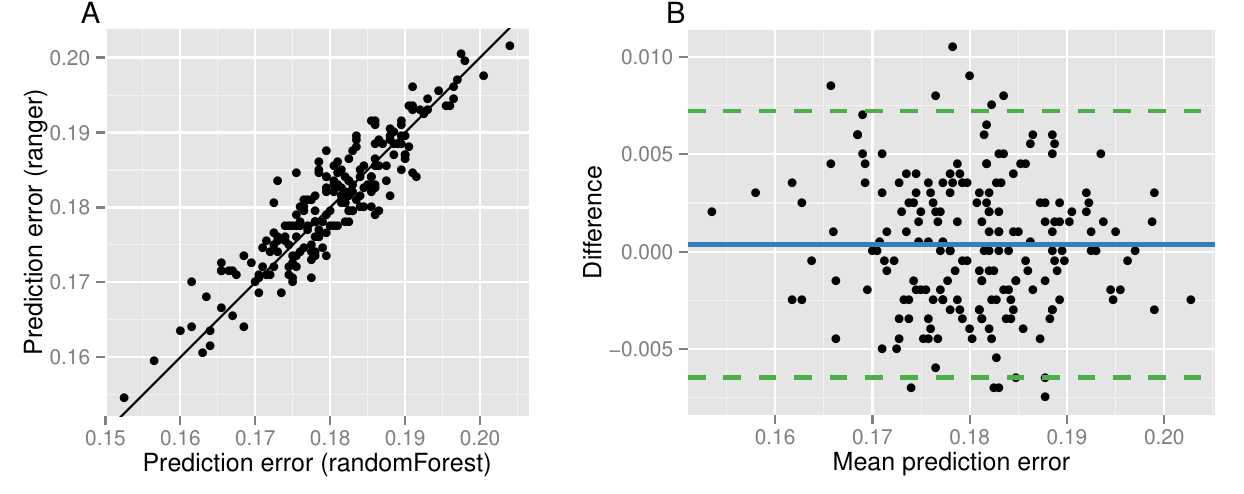}
  \caption{Validation of \pkg{ranger} from the simulation study for classification forests. Correlation of out-of-bag prediction error from \pkg{ranger} and the \pkg{randomForest} package as a scatter plot (A) and a Bland-Altman plot (B). In B, the solid line represents the mean difference and the dashed lines $\pm$\,1.96\,SD of the difference.}
  \label{fig:validationError}
\end{figure}

First, data was simulated for a dichotomous endpoint with 2000 samples and 50 features. A logit model was used to simulate an effect for 5 features and no effect for the other 45 features. We generated 200 of these datasets and grew a forest with 5000 trees with both packages on each dataset. We then compared the out-of-bag prediction errors of the packages for each dataset. The results are shown in \autoref{fig:validationError} in a scatter plot and Bland-Altman plot. No systematic difference between the two packages could be observed. We repeated the simulation for regression forests with similar results; see \autoref{fig:validationErrorRegression} for details.

\begin{figure}[htbp]
  \centering
  \includegraphics[width=1\textwidth]{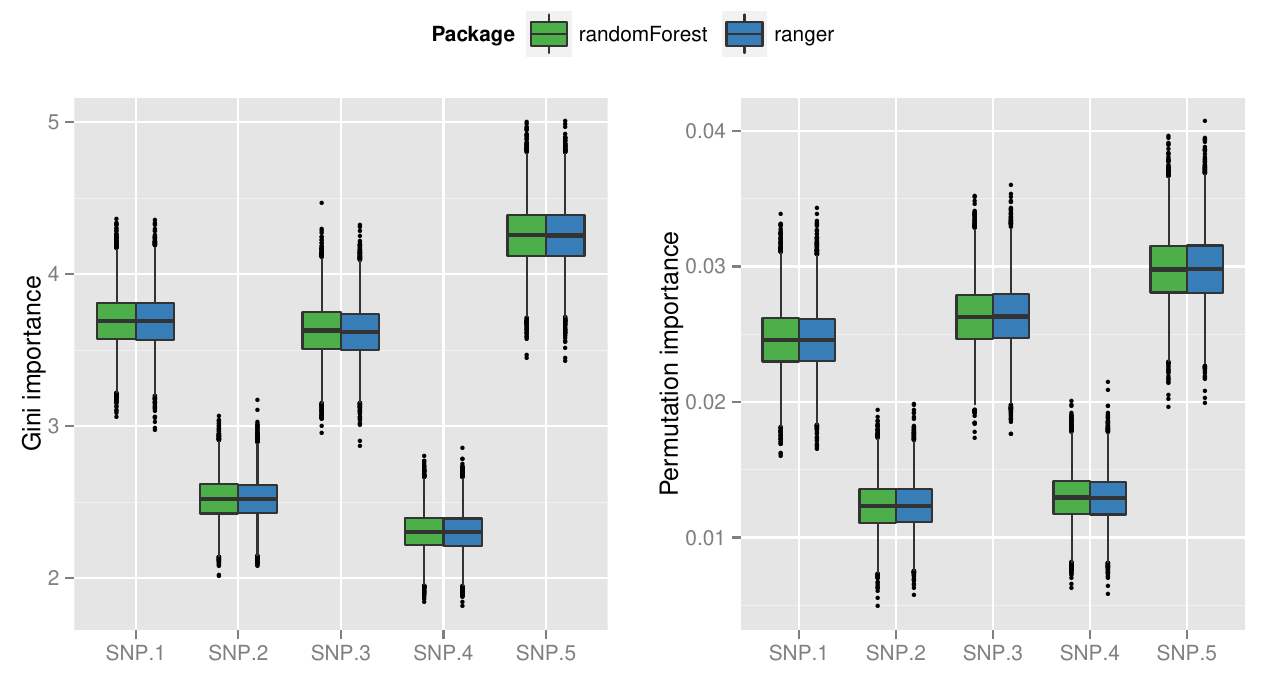}
  \caption{Validation of \pkg{ranger} from the simulation study for classification forests. Boxplots (median, quartiles, and largest non outliers) are displayed for \pkg{ranger} and the \pkg{randomForest} \proglang{R} package. Left: Gini importance for the 5 variables simulated with non-zero effect. Right: permutation importance for the same variables in the same data.}
  \label{fig:validationClassification}
\end{figure}

To compare variable importance, data was again simulated for a dichotomous endpoint with a logit model, 5 effect features and 45 noise features. Here, we grew 10,000 random forests with 500 trees each with the node impurity as split criterion and computed both the Gini importance and the permutation importance. The simulation results are provided for both importance measures and the variables with non-zero effect in \autoref{fig:validationClassification}. The Gini importance and the permutation importance results are very similar for both packages. Again, we repeated the simulation for regression forests, and results were similar; see \autoref{fig:validationRegression} for details.

\section{Runtime and memory usage}
\label{sec:runtime}

To assess the performance of the available packages for random forests in a high-dimensional setting, we compared the runtime with simulated data. First, the \proglang{R} packages \pkg{randomForest} \citep{liaw2002}, \pkg{randomForestSRC} \citep{ishwaran2015} and \pkg{Rborist} \citep{seligman2015}, the \proglang{C++} application \pkg{Random Jungle} \citep{schwarz2010, kruppa2014appl}, and the \proglang{R} version of the new implementation \pkg{ranger} were run with small simulated datasets, a varying number of features $p$, sample size $n$, number of features tried for splitting (\code{mtry}) and a varying number of trees grown in the RF. In each case, the other three parameters were kept fixed to 500 trees, 1000 samples, 1000 features and \code{mtry} $=\sqrt{p}$. The datasets mimic genetic data, consisting of $p$ single nucleotide polymorphisms (SNPs) measured on $n$ subjects. Each SNP is coded with the minor allele count, i.e., 0, 1 or 2. All analyses were run for dichotomous endpoints (classification) and continuous endpoints (regression). For classification, trees were grown until purity, and for regression the growing was stopped at a node size of 25. The simulations were repeated 20 times, and in each run all packages were used with the same data and options. For comparison, in this analysis, all benchmarks were run using a single core, and no GWAS optimized modes were used. Importance measures were disabled and recommendations for large datasets were followed. In \pkg{ranger} and \pkg{randomForest} the \code{formula} interface was not used as suggested by the help files. All other settings were set to the same values across all implementations. 

\begin{figure}[!ht]
  \centering
  \includegraphics[width=0.85\textwidth]{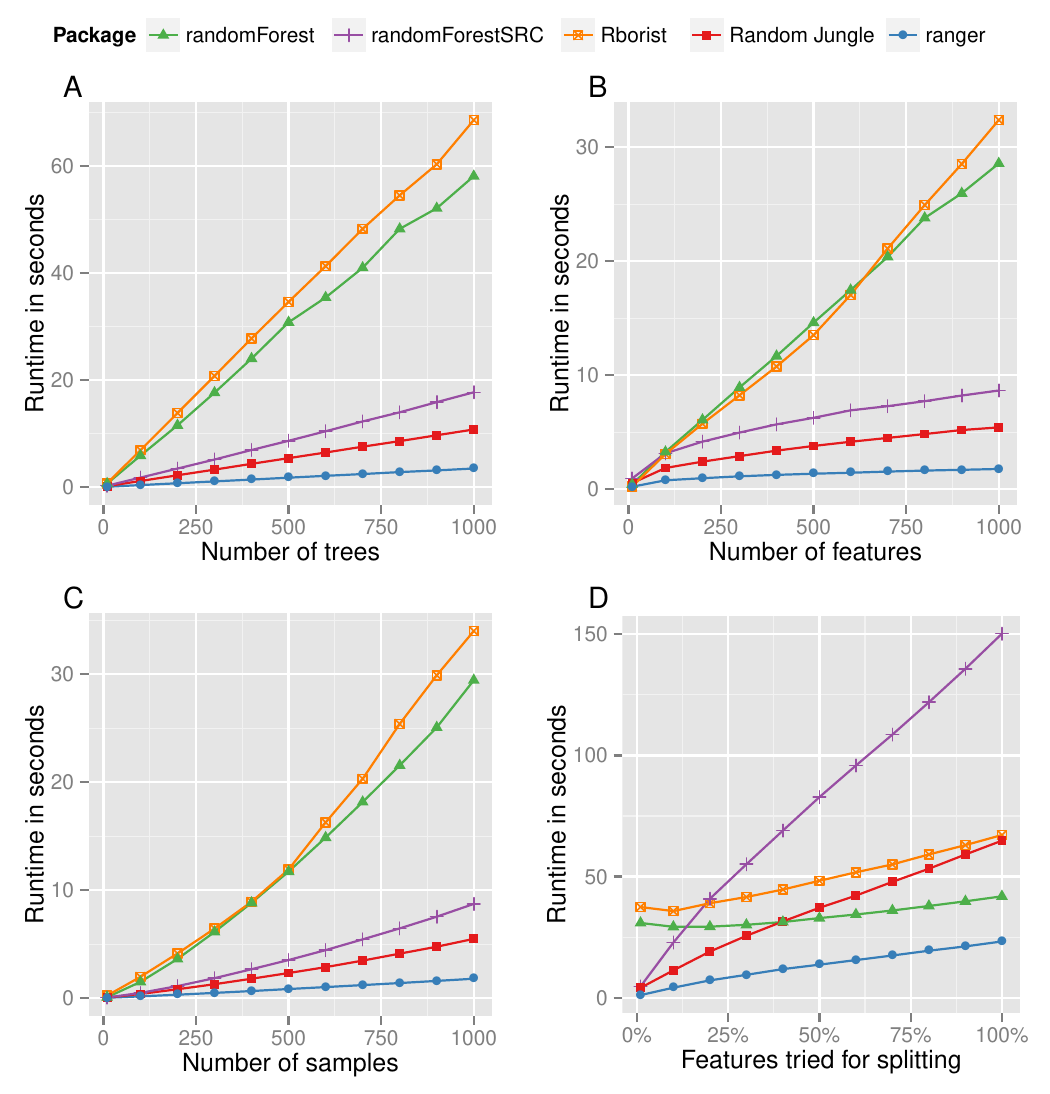}
  \caption{Runtime analysis ($y$-axis) for classification of small simulated datasets with \pkg{randomForest}, \pkg{randomForestSRC}, \pkg{Rborist}, \pkg{Random Jungle} and the new software \pkg{ranger}. A) variation of the number of trees, B) variation of the number of features, C) variation of the number of samples, D) variation of the percentage of features tried for splitting (mtry value). Each runtime corresponds to the growing of one forest. All results averaged over 20 simulation repeats.}
  \label{fig:runtimeClassification}
\end{figure}

\autoref{fig:runtimeClassification} shows the runtimes of the 5 RF packages with varying number of trees, features, samples and features tried at each split (\code{mtry}) for classification random forests. In all simulated scenarios but for high \code{mtry} values, \pkg{randomForestSRC} was faster than \pkg{randomForest}. The \pkg{Rborist} package performed similar to \pkg{randomForest}, except for high \code{mtry} values. Here, \pkg{Rborist} was slower. \pkg{Random Jungle}, in turn, was faster than \pkg{randomForestSRC} in all cases. In all simulated scenarios, \pkg{ranger} outperformed the other four packages. Runtime differences increased with the number of trees, the number of features and the sample size. All packages scaled linearly with the number of trees. When the number of features was increased, \pkg{randomForest} and \pkg{Rborist} scaled linearly, while the other packages scaled sublinearly. For the sample size, all packages scaled superlinearly. Only \pkg{ranger} scaled almost linearly. Finally, for increasing \code{mtry} values, the scaling was different. For 1$\pct$ \code{mtry}, \pkg{randomForest} and \pkg{Rborist} were slower than for 10$\pct$, with a linear increase for larger values. The packages performed with widely differing slopes: \pkg{randomForest} and \pkg{Rborist} computed slower than \pkg{randomForestSRC} for small \code{mtry} values and faster for large \code{mtry} values. \pkg{Random Jungle} was faster than both packages for small \code{mtry} values and approximately equal to \pkg{Rborist} for 100$\pct$ \code{mtry}, while \pkg{ranger} outperformed the other packages for all \code{mtry} values. Generally, although in some cases runtimes scaled in the same dimension, the differences increased considerably for larger numbers of trees, features or samples. If two or more parameters were increased simultaneously, differences added up (results not shown). 

For regression results, see \autoref{fig:runtimeRegression}. Runtimes were generally lower than for classification. In all cases except for high \code{mtry} values, \pkg{Rborist} was slowest, \pkg{randomForest}, \pkg{randomForestSRC} and \pkg{Random Jungle} about equal and \pkg{ranger} fastest. For high \code{mtry} values, \pkg{Rborist} scaled better than the other packages, but \pkg{ranger} was again fastest in all simulated scenarios.

To illustrate the impact of runtime differences on real-world applications of RF to genetic data, we performed a second simulation study. Data was simulated as in the previous study but with 10,000 subjects and 150,000 features. This is a realistic scenario for GWAS. RF were grown with 1000 trees, and \code{mtry} was set to 5000 features. The analyses were repeated for \code{mtry} values of 15,000 and 135,000 features. In addition, the maximal memory usage during runtime of the packages was recorded (\autoref{tab:memoryRuntime}). Since the \pkg{randomForest} package does not support multithreading by itself, two multicore approaches were compared. First, a simple \code{mclapply()} and \code{combine()} approach, called \pkg{randomForest} (MC), see \autoref{apx:code} for example code. Second, the package \pkg{bigrf} was used. In all simulations \pkg{randomForestSRC}, \pkg{Random Jungle} and \pkg{ranger} were run using 12 CPU cores. For \pkg{randomForest} (MC), the memory usage increased with a higher number of cores, limiting the number of cores to 3 for \code{mtry} = 5,000 and to 2 for \code{mtry} = 15,000 and \code{mtry} = 135,000. \pkg{Rborist} and \pkg{bigrf} were tried with 1 and 12 cores and \pkg{bigrf} with and without disk caching.  If possible, packages were run with data in matrix format. In \pkg{Random Jungle} the sparse mode for GWAS data was used, \pkg{ranger} was run once in standard mode, with the option \code{save.memory} enabled and in GWAS mode. 

\begin{table}[htbp]
\centering
\begin{tabular}{lrrrr}
  \hline 
 Package & \multicolumn{3}{c}{Runtime in hours} & Memory usage in GB \\ 
 & \small\code{mtry}=5000 & \small\code{mtry}=15,000 & \small\code{mtry}=135,000 & \\
  \hline
  \pkg{randomForest} & 101.24 & 116.15 & 248.60 & 39.05 \\
  \pkg{randomForest} (MC) & 32.10 & 53.84 & 110.85 & 105.77 \\
  \pkg{bigrf} & NA & NA & NA & NA \\
  \pkg{randomForestSRC} & 1.27 & 3.16 & 14.55 & 46.82 \\
  \pkg{Random Jungle} & 1.51 & 3.60 & 12.83 & 0.40 \\
  \pkg{Rborist} & NA & NA & NA & >128 \\
  \pkg{ranger} & 0.56 & 1.05 & 4.58 & 11.26 \\
  \pkg{ranger} (\code{save.memory}) & 0.93 & 2.39 & 11.15 & 0.24 \\
  \pkg{ranger} (GWAS mode) & 0.23 & 0.51 & 2.32 & 0.23 \\
   \hline
\end{tabular}
\caption{Runtime and memory usage for the analysis of a simulated dataset mimicking a genome-wide association study (GWAS) with \pkg{randomForest}, \pkg{randomForest} (MC), \pkg{bigRF}, \pkg{randomForestSRC}, \pkg{Random Jungle}, \pkg{Rborist} and the new software \pkg{ranger} in standard, memory optimized and GWAS mode. NA values indicate unsuccessful analyses: \pkg{Rborist} and \pkg{bigrf} without disk caching failed because of memory shortage for all \code{mtry} values and number of CPU cores. With disk caching, we stopped \pkg{bigrf} after 16 days of computation.}
\label{tab:memoryRuntime}
\end{table}

The \pkg{Rborist} and \pkg{bigrf} without disk caching were unable to handle the dataset. Memory usage steadily grew in the tree growing process. After several hours the system finally terminated the process because of too high memory usage, and no runtime could be measured. With disk caching, we stopped \pkg{bigrf} after 16 days of computation. All other packages successfully completed the analysis. For all values of \code{mtry}, \pkg{randomForest} used about 39 Gigabyte of system memory, and it took more than 100 hours to finish. With the \code{mclapply()} and \code{combine()} approach, runtime was reduced, but memory usage increased. Interestingly, super-linear parallel scaling was achieved in all cases. \pkg{Random Jungle} and \pkg{randomForestSRC} achieved a considerable speedup, with very low memory usage of \pkg{Random Jungle} and high memory usage of \pkg{randomForestSRC}. Finally, \pkg{ranger} was fastest in all three modes. With the \code{save.memory} option, memory usage was reduced but runtime increased, and with the GWAS mode, runtime and memory usage were both lowest of all compared packages. 

In the previous benchmarks, high dimensional data was analyzed. To compare the implementations for low dimensional data, we performed a simulation study with datasets containing 100,000 samples and 100 features. Since performance of most packages varies with the number of unique values per features, we performed all benchmarks for dichotomous and continuous features. We grew 1000 classification trees per random forest with each package. All packages but \pkg{randomForest}, were run using 12 CPU cores. All tuning parameters were set to the same values as in the previous benchmarks. \pkg{Random Jungle} was run in non-GWAS mode, \pkg{bigrf} with and without disk caching and \pkg{ranger} in standard mode. Again, maximal memory usage during runtime was measured (\autoref{tab:memoryRuntimeLowDim}).

\begin{table}[htbp]
\centering
\begin{tabular}{lrrrr}
   \hline 
  Package & \multicolumn{2}{c}{Runtime in minutes} & Memory usage in GB \\ 
  & dichotomous features & cont. features \\
   \hline
   \pkg{randomForest} & 25.88 & 34.78 & 7.76 \\
   \pkg{randomForest} (MC) & 2.98 & 3.75 & 9.17 \\
   \pkg{bigrf} (memory) & 5.22 & 5.72 & 11.86 \\
   \pkg{bigrf} (disk) & 24.46 & 26.39 & 11.33  \\
   \pkg{randomForestSRC} & 8.94 & 9.45 & 8.85\\
   \pkg{Random Jungle} & 0.87 & 1367.61 & 1.01 \\
   \pkg{Rborist} & 1.40 & 2.13 & 0.84 \\
   \pkg{ranger} & 0.69 & 5.49 & 3.11 \\
    \hline
\end{tabular}
\caption{Runtime and memory usage for the analysis of a simulated dataset with 100,000 samples and 100 features with \pkg{randomForest}, \pkg{randomForest} (MC), \pkg{bigRF} in memory and disk caching mode, \pkg{randomForestSRC}, \pkg{Random Jungle}, \pkg{Rborist} and the new software \pkg{ranger}. In each run, 1000 classification trees were grown.}
\label{tab:memoryRuntimeLowDim}
\end{table}

Runtime and memory usage varied widely between packages and between dichotomous and continuous features in some cases. For this kind of dataset \pkg{randomForest} (MC) could be run with 12 cores and it achieved a considerable speedup compared to \pkg{randomForest}, while the memory usage was only slightly higher. The in-memory \pkg{bigrf} version was faster than \pkg{randomForest} but slower than \pkg{randomForest} (MC), while the disk caching version was comparatively slow. Memory usage was only slightly reduced with disk caching. The \pkg{randomForestSRC} package was slower than \pkg{randomForest} (MC) and the memory usage approximately equal. \pkg{RandomJungle} was very fast for dichotomous features but slow for continuous features, with low memory usage for both. In contrast, \pkg{Rborist} was very fast and memory efficient for dichotomous and continuous features. Finally, \pkg{ranger} was the fastest implementation for dichotomous features, but was outperformed by \pkg{Rborist} for continuous features. Memory usage was lower in \pkg{ranger} than in \pkg{randomForest} or \pkg{bigrf}, but higher than in \pkg{Rborist} or \pkg{Random Jungle}. 

\begin{figure}[htbp]
  \centering
  \includegraphics[width=0.57\textwidth]{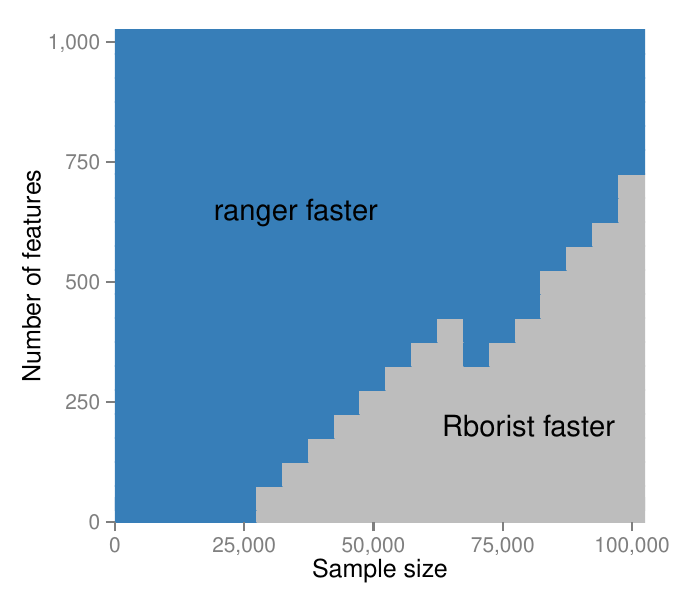}
  \caption{Runtime comparison of classification forests with \pkg{ranger} and \pkg{Rborist} for continuous features and 441 combinations of sample sizes between 10 and 100,000 and 10 to 1000 features. Regions where \pkg{ranger} was faster are marked in blue and regions where \pkg{Rborist} was faster in grey. In each run, 1000 trees were grown using 12 CPU cores.}
  \label{fig:runtime_matrix_rborist_cont}
\end{figure}

The results from Tables \ref{tab:memoryRuntime} and \ref{tab:memoryRuntimeLowDim} show that \pkg{ranger} is the fastest implementation for features with few unique values in case of high and low dimensional data. However, \pkg{Rborist} is faster for continuous features and low dimensional data with large sample sizes. To find the fastest implementation in all cases, we compared \pkg{Rborist} and \pkg{ranger} for continuous features with varying samples sizes and numbers of features. \autoref{fig:runtime_matrix_rborist_cont} shows the results of comparisons for 441 combinations of sample sizes between 10 and 100,000 and 10 to 1000 features. For each combination, a blue rectangle was drawn if \pkg{ranger} was faster or a grey rectangle if \pkg{Rborist} was faster. For sample sizes below 25,000 \pkg{ranger} was faster for all numbers of features, but for larger sample sizes a threshold was observed: \pkg{Rborist} was faster for few features and slower for many features. With increasing sample size the threshold increased approximately linearly. We observed that \pkg{ranger} scaled slightly better with the number of CPU cores used, and thus for other values the threshold can vary. 

The present simulation studies show that there is not one best random forest implementation for analyzing all kinds of large datasets. Most packages are optimized for specific properties of the features in the dataset. For example, \pkg{Random Jungle} is evidently optimized for GWAS data and \pkg{Rborist} for low dimensional data with very large sample sizes. To optimize runtime in applications, \pkg{Rborist} could be used for low dimensional data with large sample sizes ($n$ > 25,000) and \pkg{ranger} in all other cases. If memory is sparse, the \code{save.memory} option in \pkg{ranger} should be set, and if GWAS data is analyzed, \pkg{ranger} should be run in the optimized GWAS mode to take advantage of best runtime and memory usage at the same time. It should be noted that not all packages lead to the same results in all cases. For example, in classification trees, \pkg{Rborist} grows until the decrease of impurity in a node is below a threshold, while \pkg{randomForest} grows until purity. As a result, with the default settings, \pkg{Rborist} grows smaller trees than \pkg{randomForest}. 

\section{Conclusions}

We introduced the \proglang{C++} application \pkg{ranger} and its corresponding \proglang{R} package. Out-of-bag prediction errors and variable importance measures obtained from \pkg{ranger} and a standard implementation were very similar, thereby demonstrating the validity of our implementation. In simulation studies we demonstrated the computational and memory efficiency of \pkg{ranger}. The runtime scaled best for the number of features, samples, trees and the \code{mtry} value, and we are not aware of any faster RF implementation for high dimensional data with many features. The number of trees required for the analysis depends on several factors, such as the size of the dataset and the aim of the study. However, runtime and memory usage might also have affected the choice of the number of trees and other parameters. With faster implementations available, these computational challenges can be tackled.

\paragraph{Computational details.} A 64-bit linux platform with two Intel Xeon 2.7 GHz CPUs, 8 cores each, 128 GByte RAM and \proglang{R} 3.1.2 \citep{r2014} was used for all computations. For all analyses, the computing node was used exclusively for that task. The runtime of \pkg{ranger} 0.2.5, \pkg{Random Jungle} 2.1.0 \citep{kruppa2014appl}, \pkg{randomForest} 4.6-10 \citep{liaw2002}, \pkg{randomForestSRC} 1.6.1 \citep{ishwaran2015}, \pkg{Rborist} 0.1-0 \citep{seligman2015} and \pkg{bigrf} 0.1-11 \citep{lim2014} was compared using \pkg{microbenchmark} 1.4-2 \citep{mersmann2013}. For visualization, \pkg{ggplot2} 1.0.0  \citep{wickham2009} was used. \autoref{fig:runtime_matrix_rborist_cont} was created using \pkg{BatchJobs} 1.5 \citep{bischl2015}. For Tables \ref{tab:memoryRuntime} and \ref{tab:memoryRuntimeLowDim} all packages were run in separate sessions. Memory usage was measured using the Linux command \code{smem}. For the \proglang{R} packages, the garbage collector was called with the function \code{gc()} after data preprocessing. To obtain the reported memory usages in Tables \ref{tab:memoryRuntime} and \ref{tab:memoryRuntimeLowDim}, the memory usage after the \code{gc()} call was subtracted from the maximum usage during RF analysis.

\section*{Acknowledgments}
This work was supported in part by the European Union (HEALTH-2011-278913) and by the DFG Cluster of Excellence ``Inflammation at Interfaces''. The authors are grateful to the editor and two anonymous referees for helpful comments and to Inke R.\ K{\"o}nig and Silke Szymczak for valuable discussions.

\bibliography{jss1445}

\clearpage
\begin{appendix}
  
  \section{Additional figures}\label{apx:figures}
  \begin{figure}[h!]
    \centering
    \includegraphics[width=1\textwidth]{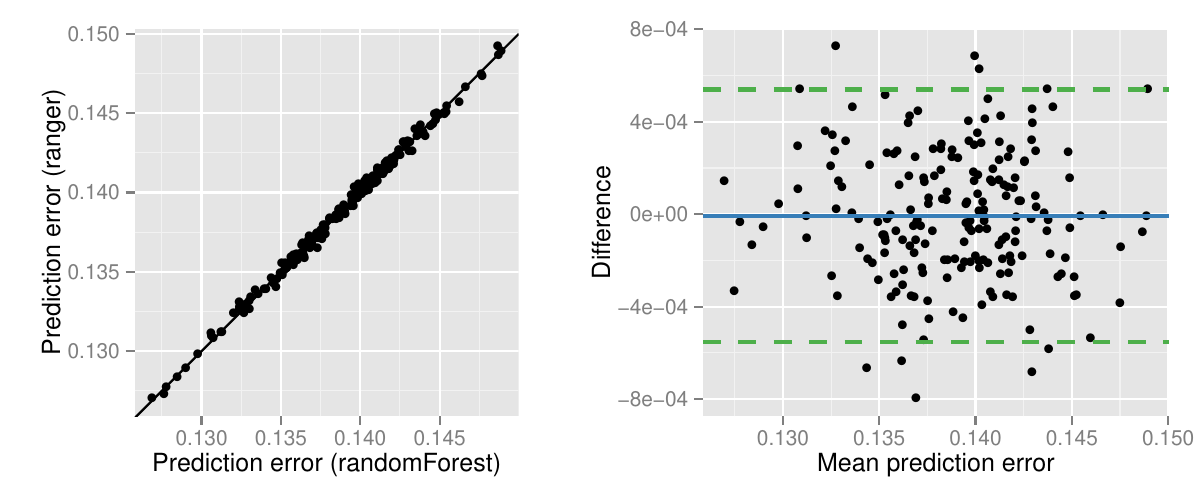}
    \caption{Validation of \pkg{ranger} from the simulation study for regression forests. Correlation of out-of-bag prediction error from \pkg{ranger} and the \pkg{randomForest} package as a scatter plot (A) and a Bland-Altman plot (B). In B, the solid line represents the mean difference and the dashed lines $\pm$\,1.96\,SD of the difference.}
    \label{fig:validationErrorRegression}
  \end{figure}
  
  \begin{figure}[h!]
    \centering
    \includegraphics[width=1\textwidth]{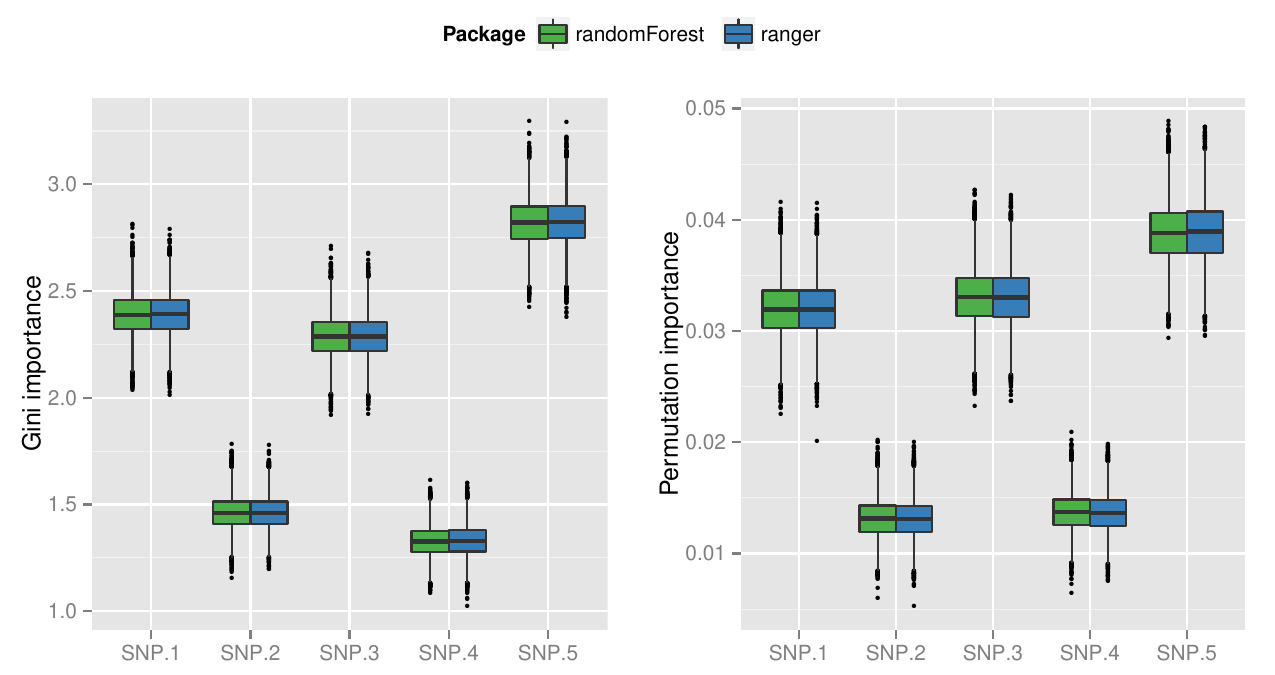}
    \caption{Validation of \pkg{ranger} from the simulation study for regression forests. Boxplots (median, quartiles, and largest non outliers) are displayed for \pkg{ranger} and the \pkg{randomForest} \proglang{R} package. Left: Gini importance for the 5 variables simulated with non-zero effect. Right: permutation importance for the same variables in the same data.}
    \label{fig:validationRegression}
  \end{figure}
  
  \begin{figure}[htbp]
    \centering
    \includegraphics[width=0.85\textwidth]{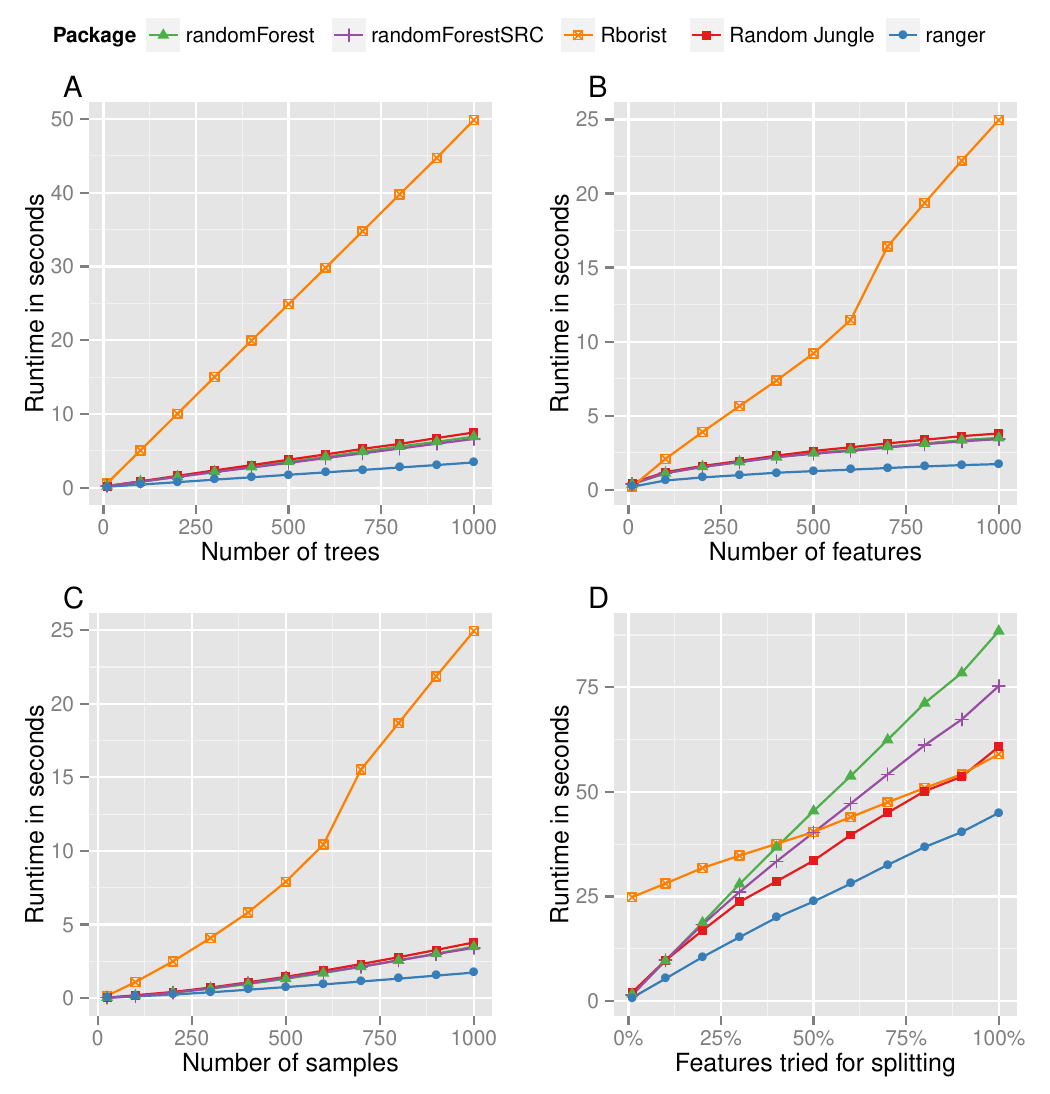}
    \caption{Runtime analysis ($y$-axis) for regression of small simulated datasets with \pkg{randomForest}, \pkg{randomForestSRC}, \pkg{Rborist}, \pkg{Random Jungle} and the new software \pkg{ranger}. A) variation of the number of trees, B) variation of the number of features, C) variation of the number of samples, D) variation of the percentage of features tried for splitting (mtry value). Each runtime corresponds to the growing of one forest. All results averaged over 20 simulation repeats.}
    \label{fig:runtimeRegression}
  \end{figure}
  
  \section[Multicore randomForest code]{Multicore \pkg{randomForest} code}\label{apx:code}
  \begin{Code}
    mcrf <- function(y, x, ntree, ncore, ...) {
      ntrees <- rep(ntree
                                              rep(1, ntree
      rfs <- mclapply(ntrees, function(n) {
        randomForest(x = x, y = y, ntree = n, ...)
      }, mc.cores = ncore)
      do.call(combine, rfs)
    }
  \end{Code}
\end{appendix}


\end{document}